\documentclass[letterpaper, 10 pt, conference]{ieeeconf}  

\IEEEoverridecommandlockouts                              

\usepackage{amsmath} 
\usepackage{amssymb}  
\usepackage{graphicx}
\usepackage{balance}
\usepackage{lipsum}
\usepackage{xcolor}
\usepackage{hyperref}
\usepackage{cleveref}
\usepackage{amsmath} 
\usepackage{amssymb}  
\usepackage{esvect}
\usepackage{algorithm,algpseudocode}
\usepackage{amssymb} 
\usepackage{multirow}
\usepackage{float}
\usepackage{adjustbox}
\usepackage{booktabs}

\DeclareMathOperator*{\argmin}{arg\,min}
\DeclareMathOperator*{\argmax}{arg\,max}

\definecolor{nscolor}{rgb}{0.188, 0.478, 0.624}

\algrenewcommand{\algorithmiccomment}[1]{// #1}

\newcommand{\objectives}{{\mathcal{E}}}

\newcommand{\control}{\Delta \mathbf{p}_{t}}
\newcommand{\ring}{\mathcal{R}}
\newcommand{\dissectionline}{D}
\newcommand{\dissectionpointcoor}{k}

\newcommand{\methodname}{MEDiC}
\newcommand{\methodnamespace}{MEDiC }
\newcommand{\methodfullname}{\textbf{M}aximizing \textbf{E}xposure for \textbf{Di}ssection and \textbf{C}autery}

\title{\LARGE
\textbf{\methodname}: Autonomous Surgical Robotic Assistance to \methodfullname
}

\author{Xiao Liang$^*{}^1$, Chung-Pang Wang$^*{}^1$, Nikhil Uday Shinde$^1$, Fei Liu$^2$, Florian Richter$^1$, Michael Yip$^1$ 
\thanks{$^*$These authors contributed equally to this paper.}
\thanks{This project was funded by the US Army Telemedicine and Advanced Technologies Research Center (TATRC), NIH \#1R01CA278703-01, and NSF Career Award \#2045803}
\thanks{$^{1}$ Department of Electrical and Computer Engineering, University of California San Diego, La Jolla, CA 92093, USA {\tt\small \{x5liang, chw120, nshinde, frichter, yip\}@ucsd.edu}}%
\thanks{$^{2}$ Department of Electrical Engineering \& Computer Science, University of Tennessee, Knoxville, TN 37996, USA {\tt\small \{fliu33\}@utk.edu}}%
}

\begin{document}

\maketitle
\thispagestyle{empty}
\pagestyle{empty}

\begin{abstract}

    Surgical automation has the capability to improve the consistency of patient outcomes and broaden access to advanced surgical care in underprivileged communities. 
    Shared autonomy, where the robot automates routine subtasks while the surgeon retains partial teleoperative control, offers great potential to make an impact. 
    In this paper we focus on one important skill within surgical shared autonomy: Automating robotic assistance to maximize visual exposure and apply tissue tension for dissection and cautery. 
    Ensuring consistent exposure to visualize the surgical site is crucial for both efficiency and patient safety. 
    However, achieving this is highly challenging due to the complexities of manipulating deformable volumetric tissues that are prevalent in surgery.
    To address these challenges we propose \methodname, a framework for autonomous surgical robotic assistance to \methodfullname.
    We integrate a differentiable physics model with perceptual feedback to achieve our two key objectives: 1) Maximizing tissue exposure and applying tension for a specified dissection site through visual-servoing conrol and 2) Selecting optimal control positions for a dissection target based on deformable Jacobian analysis. 
    We quantitatively assess our method through repeated real robot experiments on a tissue phantom, and showcase its capabilities through dissection experiments using shared autonomy on real animal tissue.

\end{abstract}


\section{INTRODUCTION}
Surgical automation has the potential to alleviate strain on healthcare systems, improve the consistency of patient outcomes, and broaden access to advanced surgical care in underserved communities. 
Advancements in shared autonomy and automation of routine surgical subtasks can reduce surgeon fatigue, and open the door to full automation of routine procedures, extending care to remote and underprivileged populations.
The introduction of open-source platforms like the da Vinci Research Kit \cite{dvrk} and advancements in deformable simulators \cite{liang2023real} have significantly propelled research in surgical automation.
Previous works have demonstrated progress in the automation of tasks like deformable tissue manipulation \cite{liang2023real, shinde2024jiggle, Fei_2021_ICRA}, hemostasis \cite{Jingbin_2021, Florian_2021_RAL}, suture needle handoff \cite{dettorre2018automatedpickupsuturingneedles, shinde2024surestep}, suturing \cite{chiu2023real, Neelay_2023_ICRA, joglekar2024autonomousimagetograsproboticsuturing}, tissue dissection \cite{oh2023framework}, and tissue retraction \cite{nguyen2019manipulating, shahkoo2023autonomous, 7989275, 7139344}.
\begin{figure}[t]
    \centering
    \includegraphics[width=0.97\linewidth]{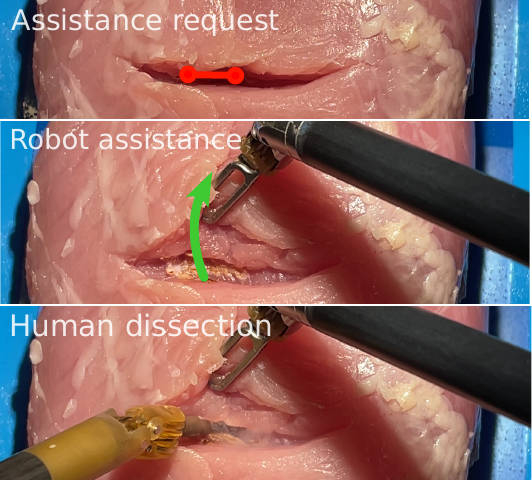} 
    \caption{Autonomous assistance for tissue dissection in robotic surgery. In response to a assistance request from a human operator, a robotic gripper improves visibility of the dissection site by pushing and stretching the tissue. The gripper's action is informed by a visual servoing controller. The human operator dissects the tissue after the dissection site is exposed.
    }
    \vspace{-1.5em}
    \label{fig:cover_photo}
\end{figure}

In this paper we focus on the surgical subtask of Maximizing Exposure (ME) for Dissection and Cautery (DiC). 
Specifically, we examine shared autonomy scenarios where the robot assists the surgeon with dissection tasks.
Maintaining adequate exposure to consistently visualize the surgical site is emphasized as a fundamental aspect of good surgical dissection technique \cite{FundamentalsOfDissection}.
Proper operative exposure is crucial to ensure both efficiency and patient safety in surgery \cite{OperativeTechniquesCardiac}. 
For tasks like dissection, this involves an assistive tool deforming the surrounding tissue to maximize exposure around the dissection site while also applying appropriate tension \cite{elek2017towards} to facilitate a successful cut. 
For common procedures like the resection of tumors from organs, this involves manipulating complex, three dimensional, deformable geometries.

Previous work has shown success in automating tissue retraction. 
\cite{nguyen2019manipulating, shahkoo2023autonomous, 7989275, 7139344} focus on both initial tool positioning and control policies to aid the cutting of planar objects, highlighting the importance of both components for successful dissection assistance. 
\cite{McKinley2015AutonomousMS, ge2021supervised, 10354422} demonstrate advancements in tissue retraction automation, however, these methods largely employ fixed tool positioning or simple control policies resulting in suboptimal performance. 
While these approaches yield some promising results, they fail to explicitly consider the important objective of maximizing exposure of the dissection site, instead primarily focusing on maintaining tension. 
Additionally, these methods only address the problem for thin-shell tissues, completely overlooking the challenges of complex, deformable, volumetric geometries, which is crucial for real-world surgeries.


In this work we propose \methodname: a method for Autonomous Surgical Robotic Assistance to \methodfullname.
Our method extends beyond thin-shell geometries to handle complex volumetric deformable environments. 
We utilize a differentiable physics-based simulation combined with real-time visual feedback from a stereo endoscopic camera to model three dimensional tissues. 
The key technical contributions of this paper are: 
\begin{itemize}
    \item We propose a visual-servoing control method using the Jacobian of our differentiable physics model and visual feedback, to maximize exposure for improved visibility and apply tension for a specified dissection target.
    \item We propose a method for optimal control point selection based on deformable Jacobian analysis for efficiency and safety given a dissection target.  
    \item We quantitatively assess our method through multiple real-world robotic experiments on a tissue phantom.
    \item We demonstrate the robustness of MEDiC through a real tissue experiment. 
    We deploy our method in a shared autonomy scenario with a human-teleoperated dissector to assist in a tissue dissection task. 
\end{itemize}

\section{RELATED WORKS}
\subsection{Autonomous Dissection}
Autonomous tissue dissection is extensively explored by previous works. 
\cite{8884158, 10.1007/978-3-030-32239-7_36} developed supervised autonomous controllers for precise 3D incisions on porcine tissues. 
\cite{oh2023framework} proposed a framework for tissue boundary detection and tool tracking to enable autonomous dissection.
\cite{ge2021supervised} introduced a customized vacuum grasping system and depth estimation for full tumor resection.
The Autonomous System for Tumor Resection (ASTR) \cite{10354422}, integrates control and planning for autonomous pseudotumor resections on porcine tongues.
Another recent approach, \cite{karimi2024reward}, adopted a reinforcement learning to automate an electro-cautery dissection task from a point cloud.
Other works focus on sensing \cite{marahrens2024ultrasound} and reconstruction \cite{franccois2021image} of the dissection site.
Most existing approaches address the problem in unrealistic surgical environments, relying on direct visibility of the dissection sites.
They fail to translate effectively to real surgeries where the surgical site is obscured by organs or self occlusion from the tissue. 
Previous works try to manage these challenges using simple policies to increase exposure, but fail to generalize to real world scenarios. 
To solve this problem, we propose a framework to assist a dissection agent to increase the exposure of a dissection site. 
This work can potentially complement previous works that focus solely on dissection.


\subsection{Robotic Dissection Assistance Strategies}
Previous work addresses the dissection assistance mainly in two directions: (1) tensioning strategies to deform soft material to a tensioning configuration, and (2) tissue retraction techniques to reveal hidden surgical sites. 
\cite{DBLP:journals/corr/abs-1901-03327, nguyen2019manipulating, shahkoo2023autonomous, 7989275} explored the problem of learning a tensioning policy using deep reinforcement learning. 
Their frameworks are equipped with the ability to search for best grasping point to perform tensioning action, and control a dissection tool for cutting. 
However, their work is mostly limited to planar thin-shell deformable objects. 
There are also previous attempts on improving surgical site exposure. 
\cite{5354075, patil2010toward} investigated retraction motion planning and optimization for the tissue gripper based on physics tissue model. 
\cite{shinde2024jiggle} proposed an active sensing framework that identify boundary attachment in a surgical environment for finding potential dissection region. 
\cite{meli2021autonomous} incorporated task-level reasoning with a tissue simulation to improve planning efficiency for thin-shell tissue retraction. 
Most of these works are limited to retracting thin-shell tissues and overlook the challenges in manipulating volumetric deformables. 
Additionally these methods fail to explicitly consider the metric of maximizing exposure of the surgical site, instead often solely focusing on objectives like tissue tensioning. 
By providing a methodology to maximize exposure when dissecting deformable volumetric geometries, \methodnamespace makes a substantial contribution toward advancing autonomous robotic assistance for dissection and cautery.

\section{PRELIMINARIES}
\label{sec:xpbd}
In this work we choose a differentiable quasi-static Extended Position-based Dynamics (XPBD) to model deformable volumetric tissues. 
Let $\mathbf{x}_t$ be a set of surface and internal particles representing a volumetric tissue, $M$ be a tetrahedral mesh representing connectivity of $\mathbf{x}_t$, and $M^\text{surf}:\{F_1,...,F_b\}$ be a triangle mesh that is the surface of $M$. 
$M(\mathbf{x}_t)$ or $F(\mathbf{x}_t)$ is used to denote the geometry of the mesh after $\mathbf{x}_t$ has deformed.
Let $\mathbf{p}_t$ denote a robot's end-effector (EE) position at time $t$. 
The EE first interacts with tissue at point $\mathbf{p}_0$ on the tissue surface. 
The mesh is controlled by rigidly coupling the nearest triangle face, $F_{\mathbf{p}}$, to the EE's translation.
This rigid coupling is enforced as a boundary condition. 
In the real world, this can be realized by frictional contact or grasping. 
We limit this work to the frictional contact case. 

The XPBD simulator solves for the updated state $\mathbf{x_{t}}$, after control, by minimizing the system's total potential energy:
\begin{equation}\label{eqn:pbd_objective}\small
\begin{split}
\mathbf{x}_{t} &= \argmin_{\mathbf{x}_{t}} \frac{1}{2} \mathbf{C}(\mathbf{x}_{t})^\top \mathbf{K} \mathbf{C}(\mathbf{x}_{t})  \\
\textbf{s.t.}& \quad x_{t}^i - x_{0}^i = \mathbf{p}_{t} - \mathbf{p}_0, x^i \in F_\textbf{p} 
\end{split}
\end{equation}
where $\mathbf{C}(\mathbf{x}_{t})$ is a set of geometric constraints that jointly determine the tissue's deformation behavior, and $\mathbf{K}$ stores their weights. In this work, we use distance and shape matching constraints.
We summarize a forward step of the XPBD simulator that solves Eqn.~\ref{eqn:pbd_objective} as $\mathbf{x}_{t} = f(\mathbf{x}_{t-1},  \mathbf{p}_{t})$.

\renewcommand{\algorithmiccomment}[1]{\hfill{\small\texttt{$\triangleright$\ #1}}}
\begin{algorithm}[!t]
\small
    \caption{Autonomous dissection assistance}
    \label{alg:full_algorithm}
    \begin{algorithmic}[1]
    \While {$Assistance\ needed$}
        \State $\mathbf{x}_0, M \gets meshReconstruction()$
         \State $f \gets initSimulation(\mathbf{x}_0, M)$
        \State $D \gets humanRequest()$
        \State $\mathbf{L}\gets\{\dissectionpointcoor_1,...,\dissectionpointcoor_L\}$,\ $\mathbf{R}\gets\{r_1,...r_R\}$\Comment{{\footnotesize parameters}}
        \State $O \gets initObservation(D, L, R, M)$\Comment{{\footnotesize \autoref{sec:dissection_assistance_objectives}}}
        \State $\mathbf{p}_0 \gets APS(\mathbf{x}_0, M, f)$ 
        \Comment{{\footnotesize Apply \autoref{eqn:aps}}}
        \For {$\textit{numIterations}$} 
              \State $\mathbf{z}_t, \mathbf{I}_\text{seg} \gets perception()$
              \Comment{{\footnotesize \text{stereo vision}}}
              \State $M^\text{vis} \gets rayCasting(M, \mathbf{I}_\text{seg}, \mathbf{x}_{t-1})$
              \State $\mathbf{x}_{t} \gets tissueState(f(\mathbf{x}_{t-1}, \mathbf{p}_{t}), M^\text{vis}, \mathbf{z}_t)$
            \Comment{{\footnotesize \autoref{eqn:pbd_objective}, \autoref{eqn:chamfer}}}
              \State $\control \gets ControlStep\Bigl(\objectives\bigl(O(\mathbf{x}_t)\bigr)\Bigr)$ \Comment{{\footnotesize \autoref{eqn:pdrule}}}
              \State $\mathbf{p}_{t+1} = \mathbf{p}_{t} + \control$ \Comment{{\footnotesize Move EE}}
          \EndFor
        \State $humanDissect(D)$
    \EndWhile
    \end{algorithmic}
\end{algorithm}


\section{METHODS}
Our primary goal in this work is maximizing exposure of the dissection site. 
We formulate this objective as the minimization of a visual servoing error :
\begin{equation}\label{eqn:error_function}
    \begin{split}
        \objectives(t) = o^* - O(\mathbf{z}_t, \mathbf{x}_t, \dissectionline, f) 
    \end{split}
\end{equation}
where $O(\cdot)$ is an observation function based on the current visual data $\mathbf{z}_t$, tissue state $\mathbf{x}_t$, dissection target $\dissectionline$, and physics-based priors from the simulator $f$.
$O(\cdot)$ quantifies the exposure of the dissection site near the target, with $o^*$ representing a desired exposure goal. 
To solve this problem, we leverage the Jacobian from our differentiable simulator. 
    \begin{equation}
        \begin{split}
            \mathbf{J}_d & = \frac{\partial f(\mathbf{x}_t, \mathbf{p}_t)}{\partial \mathbf{p}}\bigg|_{\mathbf{p}_t} 
        \end{split}
    \end{equation}
$\mathbf{J}_d$ is the deformation Jacobian matrix. We use a Proportional-derivative (PD) control law to minimize the error in Eqn. \ref{eqn:error_function} with the robot's EE:
    \begin{equation}\label{eqn:pdrule}
        \begin{split}
            \control  =& K_p (\mathbf{J}_O\mathbf{J}_d)^\dagger \objectives\bigl(O(\cdot)\bigr)+ K_d (\mathbf{J}_O\mathbf{J}_d)^\dagger \frac{d \objectives\bigl(O(\cdot)\bigr)}{dt}\\
        \end{split}
    \end{equation}
where $K_p, K_d$ are the gains that we tune, and $\mathbf{J}_O$ is the Jacobian of observation function $O$ w.r.t to tissue state $\mathbf{x}_t$.

To implement this control strategy, we divide our methods into 3 components: (1) design of the error and observation functions $\objectives, O$; (2) intelligent selection of the initial control point $\mathbf{p}_0$ to realize the control goal efficiently; and (3) incorporation of robot perception for tissue state estimation.
Our overall algorithm is shown in \autoref{alg:full_algorithm}.

\subsection{Dissection Assistance Error}\label{sec:dissection_assistance_objectives}
\begin{figure}[t]
    \centering
    \includegraphics[width=0.9\linewidth]{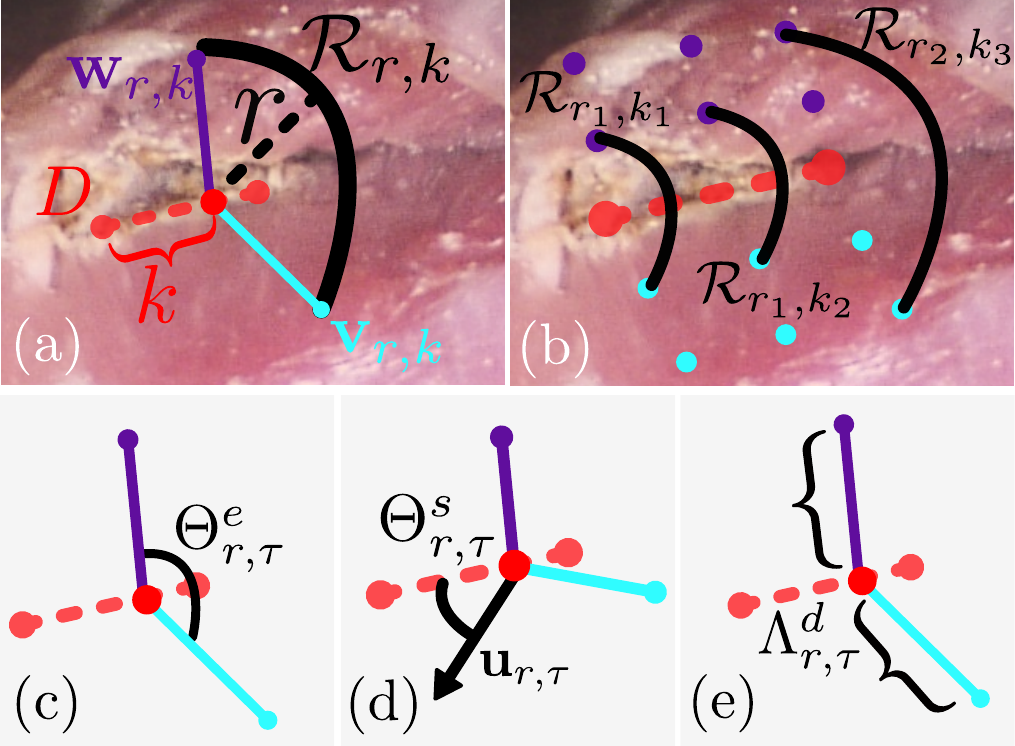} 
    \caption{Visualization of the geometrical concepts to define our  objectives. The top left figure shows how a single feature pair, $v_{r,k}, w_{r,k}$, are found given a dissection line $D$. The top right figure illustrate the idea of selecting multiple rings.
    In the bottom from left to right, they illustrate wedge expansion, shear regulation, and stretch enforcement objectives.
    }
    \vspace{-1.5em}
    \label{fig:visual_servoing_objective}
\end{figure}

Since surgical site exposure is abstract, our key idea is to use local angles and distance measurements to approximate it.
Let the dissection target $\dissectionline : \{q_1, q_2\}$ be a line segment such that $\dissectionline(\dissectionpointcoor) = q_1 + \dissectionpointcoor (q_2-q_1), \dissectionpointcoor\in[0,1]$ is any point on the dissection segment. 
Let $\ring(r, c, \mathbf{n})$ be a ring geometry in $\mathbb{R}^3$ parameterized by a radius $r$, a center $c$, and a normal vector $\mathbf{n}$.
A pair of feature vectors, $\vv{\mathbf{v}}_{r, \dissectionpointcoor}, \vv{\mathbf{w}}_{r,\dissectionpointcoor}$, is found by computing the intersection points between the current tissue surface and a 3D ring whose center, $c$, is on the dissection segment and normal, $n$, is parallel to the dissection line:
    \begin{equation}
        \begin{split}
            \mathbf{v}_{r, \dissectionpointcoor}, \mathbf{w}_{r,\dissectionpointcoor}& = M^{\text{surf}}(\mathbf{x}_t)\cap \mathcal{R}_{r, k} \left(r, \dissectionline(\dissectionpointcoor), \vv{\mathbf{d}}_t \right)\\
            \vv{\mathbf{v}}_{r, \dissectionpointcoor}, \vv{\mathbf{w}}_{r,\dissectionpointcoor}&=\mathbf{v}_{r, \dissectionpointcoor} - \dissectionline(\dissectionpointcoor),\mathbf{w}_{r, \dissectionpointcoor} - \dissectionline(\dissectionpointcoor),
        \end{split}
    \end{equation}
where $\vv{\mathbf{d}}_t = \frac{q_2-q_1}{\|q_2-q_1\|}$ is the dissection line's direction.
To improve robustness, we consider multiple vector pairs along, as well at different distances from the dissection line. 
This is done by varying combinations of the segment points $\mathbf{L}:\{\dissectionpointcoor_1,...,\dissectionpointcoor_L\}$ and ring radius $\mathbf{R}:\{r_1,...r_R\}$.
We illustrate this geometry in Fig~\ref{fig:visual_servoing_objective}.a and Fig~\ref{fig:visual_servoing_objective}.b. 

Our dissection assistance objectives $\objectives$ consist of (1) $\objectives^e$: expansion of wedge angles between feature vector pairs on both sides of dissection segment (Fig~\ref{fig:visual_servoing_objective}.c), (2) $\objectives^s$ regulation of the shearing deformation along the dissection line (Fig~\ref{fig:visual_servoing_objective}.d). 
and (3) $\objectives^d$ increase in the length of the feature vectors to apply tissue tension while cutting (Fig~\ref{fig:visual_servoing_objective}.e). 

\textbf{Wedge Expansion:} Our controller seeks to control the wedge angle between the feature point pair which can be computed as: $\cos^{-1}(\varphi_{cos}(\vv{\mathbf{v}}_{r, \dissectionpointcoor}, \vv{\mathbf{w}}_{r, \dissectionpointcoor}))$.
$\varphi_{cos}$ is the cosine similarity function.
As we are using a Jacobian based controller, we ease gradient computation by removing the arccosine in calculating our wedge angle metric: $\Theta_{r, \dissectionpointcoor}^e = \varphi_{cos}(\vv{\mathbf{v}}_{r, \dissectionpointcoor}, \vv{\mathbf{w}}_{r, \dissectionpointcoor})$.
The wedge expansion error is then computed as: 
\begin{equation}\label{eqn:wedge_expansion}
    \objectives^e_{r, \dissectionpointcoor} = o^e  - \Theta_{r, \dissectionpointcoor}^e
\end{equation}
where $o^e$ is a target cosine similarity that we tune based on the desired expansion angle. 
For example, $o^e = -1$ corresponds to a desired angle of 180 degrees. 
In this paper, we focus on outward expansion of wedge angles within the range of $0-180$ to prevent ambiguity in our metric.

 \textbf{Shear Regulation:} Shearing is quantified as the angle between the dissection segment and the vector resulting from the cross product between a feature vector pair $\vv{\mathbf{u}}_{r, \dissectionpointcoor}= \vv{\mathbf{v}}_{r, \dissectionpointcoor}\times \vv{\mathbf{w}}_{r, \dissectionpointcoor}$. The shear angle metric is $\Theta_{r, \dissectionpointcoor}^s = \varphi_{cos}( \vv{\mathbf{u}}_{r, \dissectionpointcoor}, \vv{\mathbf{d}})$. 
Our ideal control should minimize the shearing to avoid distorting the dissection target. 
Therefore, we regulate the shearing cosine similarity to $o^s=1$ so the cross product $\vv{\mathbf{u}}_{r, \dissectionpointcoor}$ is parallel to $D$: 
\begin{equation}\label{eqn:shear_regulation}
    \objectives^s_{r, \dissectionpointcoor} = o^s -\Theta_{r, \dissectionpointcoor}^s.
\end{equation}

\textbf{Stretch Enforcement}: Visual exposure can also be improved by expanding the surface area around the dissection site. 
We achieve this by introducing a distance-based objective that aims to stretch out the feature vectors: 
\begin{equation}    
    \begin{split}
        \Lambda_{r, 
        \dissectionpointcoor}^{d, \mathbf{v}} =
        \|\vv{\mathbf{v}}_{r, \dissectionpointcoor}\|,\         \Lambda_{r, 
        \dissectionpointcoor}^{d, \mathbf{w}} =
        \|\vv{\mathbf{w}}_{r, \dissectionpointcoor}\|, 
    \end{split}
\end{equation}
The distance error for a feature vector aims to stretch the surface to some multiple of its initial value: 
\begin{equation}\label{eqn:distance_regulation}
    \begin{split}
        \objectives^{d,\textbf{v}}_{r, \dissectionpointcoor} = o^{d,\textbf{v}} - \Lambda_{r, \dissectionpointcoor}^{d,\textbf{v}}, \; \; \objectives^{d,\textbf{w}}_{r, \dissectionpointcoor} = o^{d,\textbf{w}} - \Lambda_{r, \dissectionpointcoor}^{d,\textbf{w}}
    \end{split}    
\end{equation}
where $o^{d,\textbf{v}} = \gamma^{\textbf{v}}  \Lambda_{r, \dissectionpointcoor}^{d,\textbf{v}} (x_{0})$, $o^{d,\textbf{w}} = \gamma^{\textbf{w}} \Lambda_{r, \dissectionpointcoor}^{d,\textbf{w}} (x_{0})$, initial feature distances $\{\Lambda_{r, \dissectionpointcoor}^{d,\textbf{v}} (x_{0}), \Lambda_{r, \dissectionpointcoor}^{d,\textbf{w}} (x_{0})\}$ are  computed at $t=0$ and $\gamma^\textbf{v}, \gamma^\textbf{w} > 1$ are parameters that we tune. 



Our overall observation function to quantify site exposure is a concatenation of all metrics over all feature vectors:
\begin{equation}\label{eqn:observation_function}\small
    \begin{split}
        \boldsymbol{\Theta}^e &= \begin{bmatrix}\Theta^e_{r_1, \dissectionpointcoor_1} \cdots\Theta^e_{r_R, \dissectionpointcoor_1}\cdots \Theta^e_{r_R, \dissectionpointcoor_T}\end{bmatrix}^\top,\\
        \boldsymbol{\Theta}^s &= \begin{bmatrix}\Theta^s_{r_1, \dissectionpointcoor_1} \cdots\Theta^s_{r_R, \dissectionpointcoor_1}\cdots \Theta^s_{r_R, \dissectionpointcoor_T}\end{bmatrix}^\top,\\
        \boldsymbol{\Lambda}^d &= \begin{bmatrix}\Lambda^{d, \mathbf{v}}_{r_1, \dissectionpointcoor_1} \cdots \Lambda^{d, \mathbf{v}}_{r_R, \dissectionpointcoor_T},
        \Lambda^{d, \mathbf{w}}_{r_1, \dissectionpointcoor_1} \cdots \Lambda^{d, \mathbf{w}}_{r_R, \dissectionpointcoor_T}\end{bmatrix}^\top, \\
        O&= \begin{bmatrix}\boldsymbol{\Theta}^{e\top},\boldsymbol{\Theta}^{s\top},\boldsymbol{\Lambda}^{d\top}\end{bmatrix}^\top\\
    \end{split}
\end{equation}
Note that here $O$ is a function of $\mathbf{x}_t$ and $\dissectionline$, whereas in \autoref{eqn:error_function} $O$ is also a function of visual data $\mathbf{z}_t$. Incorporation of $\mathbf{z}_t$ will be detailed later in \autoref{sec:visual_servoing}. Similarly, the error function is formed by concatenating results from \autoref{eqn:wedge_expansion}, \autoref{eqn:shear_regulation} and \autoref{eqn:distance_regulation} that 
$\objectives=\begin{bmatrix}\objectives^{e}_{r_1,\dissectionpointcoor_1}\cdots,\objectives^{s}_{r_1,\dissectionpointcoor_1}\cdots, \objectives^{d, \textbf{v}}_{r_1,\dissectionpointcoor_1} \cdots , \objectives^{d, \textbf{w}}_{r_1,\dissectionpointcoor_1}\cdots\end{bmatrix}^\top$.
\subsection{Optimal Assistance Position Selection (APS)}

We propose a metric to select the optimal control point to efficiently minimize our control error. 
We want to choose a control position on the tissue surface that enables us to deform the tissue and maximize the wedge angle without increasing the shear angle near the dissection site.

Let $\mathbf{p}^{i}_{0} \in \mathbf{P}$ be a candidate assistance position. To quantify the impact of different control positions, we start by computing the Jacobian, $\mathbf{J}_{\text{avg}}$, at each prospective control point.
It models the impact of applying a control, at the chosen control point, on the wedge and shear objectives averaged across the feature point pairs of interest. 
$\mathbf{J}_{\text{avg}} = [{\mathbf{J}^{e}_{\text{avg}}}^\top, {\mathbf{J}^{s}_{\text{avg}}}^\top]^\top$ where ${\mathbf{J}^{e}_{\text{avg}}}, {\mathbf{J}^{s}_{\text{avg}}}$ are the Jacobians with respect to the wedge and shearing objectives respectively: 
\begin{equation}\small
    \begin{split}
        \mathbf{J}^e_{\text{avg}} = \frac{\text{avg}(\boldsymbol{\Theta^e})}{\partial f(\mathbf{x}_t, \mathbf{p}^{i}_{0} )}\mathbf{J}_d,\         \mathbf{J}^s_{\text{avg}} = \frac{\text{avg}(\boldsymbol{\Theta^s})}{\partial f(\mathbf{x}_t, \mathbf{p}^{i}_{0} )}\mathbf{J}_d.
    \end{split}
\end{equation}
To convert Jacobian to a useable metric, we start by taking its Singular Value Decomposition (SVD):
\begin{equation}\small
        \begin{split}
            \mathbf{J}_\text{avg} &= \mathbf{U} \cdot \Sigma \cdot \mathbf{V}^\intercal  \\
            &= \begin{bmatrix} \mathbf{u}_1 & \mathbf{u}_2 \\ \end{bmatrix} 
            \cdot \begin{bmatrix} \sigma_1 & 0 & 0 \\ 0 & \sigma_2 & 0 \\
            \end{bmatrix} \cdot 
            \begin{bmatrix}
             \mathbf{v}_1^\top \\ \mathbf{v}_2^\top
             \\ \mathbf{v}_3^\top
            \end{bmatrix}
        \end{split}
\end{equation}
where the diagonal elements of $\Sigma \in \mathbb{R}^{2 \times 3}$ are the singular values of $\mathbf{J}$ and $\sigma_1 \geq \sigma_2 \geq 0 $, $\mathbf{U} \in \mathbb{R}^{2 \times 2} $ and $\mathbf{V}^\intercal \in \mathbb{R}^{3 \times 3} $ are the left and right singular matrix respectively. 
In the SVD, the left unitary vectors, $\mathbf{u}_{1}, \mathbf{u}_{2}$, represent the basis in which chosen controls can impact the wedge and shear angle objectives. 
As our desired controls seek to maximize wedge angle while minimizing the impact on shear, our ideal candidate point has a Jacobian where $\mathbf{u}_{1}$ and $\mathbf{u}_{2}$ are close to the standard basis vectors, $\{\mathbf{b}^{w} = [1, 0]^\top, \mathbf{b}^{s} = [0, 1]^\top \}$. 
Having the left eigenvectors aligned with the standard basis enables us to better find controls to independently impact the separate objectives, suiting our goal of maximizing one while minimizing impact on the other.
The eigenvector closer to the direction of $\mathbf{b}^{w}=[1, 0]^\top$, $\mathbf{u}^{w}$, largely impacts the wedge objective while the eigenvector closer to the direction of $\mathbf{b}^{s}=[0 ,1]^\top$, $\mathbf{u}^{s}$, captures the impact on shear.
\begin{equation}\small
        \begin{split}
            \mathbf{u}^w = \argmax_{\mathbf{u}\in \{\mathbf{u}_1, \mathbf{u}_2\}} \bigl| \varphi_{cos}(\mathbf{u}, \mathbf{b}^w)\bigr|, \; \;             \mathbf{u}^s = \argmax_{\mathbf{u}\in \{\mathbf{u}_1, \mathbf{u}_2\}} \bigl| \varphi_{cos}(\mathbf{u}, \mathbf{b}^s)\bigr|
        \end{split}
\end{equation}
Additionally, as we prioritize wedge expansion we want a Jacobian with a high impact on the wedge angle.
This is quantified by high singular values, $\sigma_{w}$, in the direction of the $u_{w}$.
We combine all the above concepts into one singular quantifiable metric $\mathcal{M}(\cdot)$: 
\begin{equation}\small
    \begin{split}
        \mathcal{M}(\mathbf{p}_i, \mathbf{x}_t, D_t)& =\bigl|  \varphi_{cos}(\mathbf{u}^w, \mathbf{b}^w)\bigr|\sigma^w  - \alpha  \bigl|  \varphi_{cos}(\mathbf{u}^s, \mathbf{b}^s)\bigr|\sigma^s\\
    \end{split}
\end{equation}
where $\alpha$ is a hyper parameter that we tune and $\sigma_w, \sigma_s$ are the singular values corresponding to left eigenvectors $\mathbf{u}^{w}, \mathbf{u}^{s}$ respectively. 
This heuristic computes a score from the disparity between the wedge and shear projection length on the standard basis, weighted by their corresponding singular values. 
By basing our heuristic off the SVD of $\textbf{J}_{d}$, we capture the intrinsic properties of deformation allowing our heuristic to offer a precise representation of the tissue's geometric and physical characteristics. 

We define the assistance position candidate set $\mathbf{P}$ as all centroids of the surface mesh triangles. 
We also constrain the control position to be camera-visible and sufficiently distant, $l_{\text{min}}$, from the dissection target. 
At last, our method greedily selects the position with the highest heuristic value that satisfies these constraints:
\begin{equation}\label{eqn:aps}\small
    \begin{split}
        \mathbf{p}^{*}_{0} &= \argmax_{\mathbf{p}_i \in \mathbf{P}}\mathcal{M}(\mathbf{p}^{i}_{0} , \mathbf{x}_t, D_t),\\
        &s.t.\ \mathbf{p}^{i}_{0}  - D_t \ge l_{\text{min}},\ \mathbf{p}^{i}_{0}  \in M^{\text{vis}}(\mathbf{x}_t)\\
    \end{split}
\end{equation}
where $M^{\text{vis}}(\mathbf{x}_t)$ represents the current visible mesh surface (more details in \autoref{sec:visual_servoing}).
\begin{figure}[t]
    \centering
    \includegraphics[width=1.\linewidth]{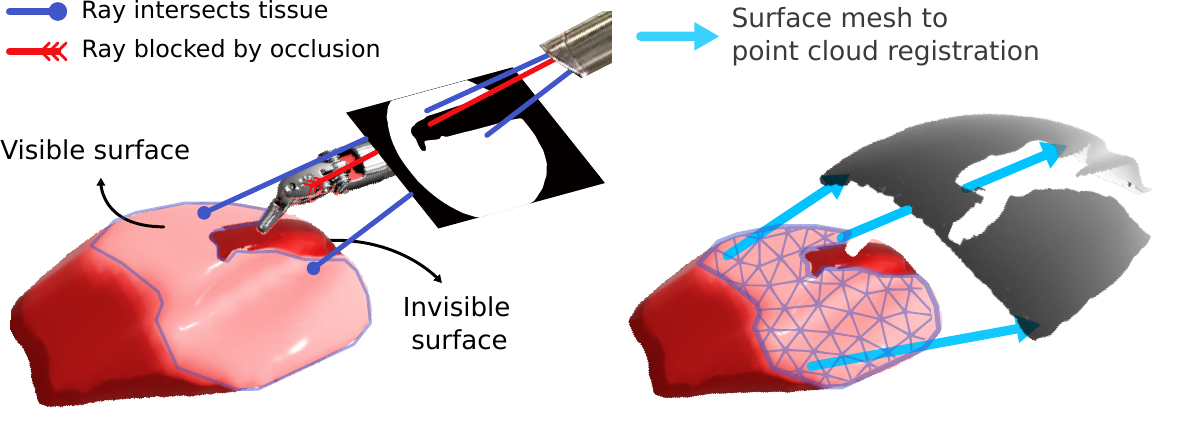}\vspace{-2.5em}
    \caption{Illustration of our tissue state estimation method. Left: Determining tissue visibility by ray-casting and checking the ray-triangle intersections.
    The circled blue region indicates the visible region of the mesh. 
    Right: Registering the volumetric mesh to the observed surface point cloud.}
    \label{fig:tissue_tracking}
    \vspace{-1em}
    \vspace{-3.5mm}
\end{figure}

\subsection{Tissue State Estimation}\label{sec:visual_servoing}
Estimation of the tissue state $\mathbf{x}_t$ is essential to provide visual feedback for control objectives.
The state of the volumetric tissue is partially observable due to self and tool-occlusion, making tissue state estimation crucial in this work.
We propose a registration method as shown in \autoref{fig:tissue_tracking} that takes the tissue visibility into account to estimate the full tissue state from partial surface point cloud. 

We determine the tissue visibility by checking whether each triangle face in the surface mesh is visible to the camera 
\begin{equation}\small
    \begin{split}        
    M^{\text{vis}}= \{F\in M^\text{surf}\ |\ \mathcal{I}\bigl(\mathbf{A}, \mathbf{I}_{\text{seg}}, F(\mathbf{x}_t)\bigr)\}
    \end{split}
\end{equation}
where $\mathcal{I}(\cdot)$ is a function that checks if any ray originated from the camera $\mathbf{A}$, intersects with the face $F$.
In addition, tissue segmentation $\mathbf{I}_\text{seg}$ is used to filter out rays that don't correspond to a pixel belong to the tissue. 

After the tissue visibility is found, a point cloud similarity objective between a surface point cloud $\mathbf{z}_t$ and the tissue's visible surface is computed using Chamfer distance as
\begin{equation}\label{eqn:chamfer}\small
    \begin{split}
        \mathcal{L}(\mathbf{x}_t, \mathbf{z}_t) =& \sum_{y\in M^\text{vis}(\mathbf{x}_t)} \min_{z\in \mathbf{z}_t}\|y - z\|^2_2 + \sum_{z\in \mathbf{z}_t} \min_{y\in M^\text{vis}(\mathbf{x}_t)}\|z -  y\|^2_2.\\
    \end{split}
\end{equation}
where $x\in M^\text{vis}(\mathbf{x}_t)$ represents random sampling point on the visible surface mesh. We incorporate the chamfer distance term as an additional objective into \autoref{eqn:pbd_objective}.
It can be then solved along side with all XPBD's geometric constraints.
The resulting tissue state will align with the point cloud observations while remaining physically plausible \cite{liang2023real}.

\section{Experiments}
We conduct extensive real-world experiments to evaluate our method's performance.
This section can be broken down into 2 parts: (1) evaluation and ablation analysis of \methodnamespace in several silicone tissue phantom scenarios where the method is tested repetitively more than 30 times, (2) an experiment with real animal tissue where  \methodnamespace assists a human operator in a series of consecutive dissection tasks.

Our real world experiments are conducted with the da Vinci Research Kit (dVRK) \cite{dvrk}.
The proposed algorithm controls one patient side manipulator (PSM) equipped with a fenestrated bipolar forceps tool, while dissection is carried out by a human teleoperator controlling a cautery spatula tool.
We keep the dVRK's stereo camera fixed during the experiments. 
We use endoscopic stereo images to perform disparity estimation \cite{li2022practical}, tissue segmentation \cite{SAM, cheng2024putting}, and reconstruction of the deformable tissue.

\begin{table}[t]
\setlength\tabcolsep{0.3em}
\centering
\caption{Comparison of our method with vs. without APS. 
Our method with APS achieves a higher exposure ratio and success rate. $n$ is the number of experiments.}
\begin{adjustbox}{width=0.45\textwidth}
    \begin{tabular}{c|lc|cc}
\toprule
&Phantom & Initial Angle & Final expansion & Success rate \\
\midrule
Ours-no-APS&P1 (n=5) &  $\Theta^w = 45^{\circ} $  & $1.29\pm0.13$ & 80\%\\
Ours-APS&P1 (n=5) &  $\Theta^w = 45^{\circ} $  & $\mathbf{1.74\pm0.17}$ & $\mathbf{100\%}$ \\
\bottomrule
\end{tabular}
\end{adjustbox}
\label{tbl:aps_table}
\end{table}
\begin{table}[t]
\setlength\tabcolsep{0.2em}
\centering
\caption{Comparison of our method with vs. without using visual servoing control. 
With visual servoing, our method achieves better results on 3 different phantom experiments. $n$ is the number of experiments.}
\begin{adjustbox}{width=0.45\textwidth}
    \begin{tabular}{c|lc|cc}
\toprule
&Phantom & Initial Angle & Final expansion & Success rate \\
\midrule
&P1 (n=3) &  $\Theta^w = 45^{\circ} $  & $1.31\pm0.33$ & 33\% \\
Ours-Hinge &P2 (n=3) & $\Theta^w = 65^{\circ} $  & $1.11\pm0.05$  & 0\% \\
&P3 (n=3)& $\Theta^w = 80^{\circ} $  &  $1.00\pm 0.11$ & 0\% \\
& Avg (n=9)&-&$1.14\pm 0.23$&11\%\\
\midrule
&P1 (n=10) &  $\Theta^w = 45^{\circ} $  & $2.07\pm0.45$ & 100\% \\
Ours&P2 (n=10) & $\Theta^w = 65^{\circ} $  & $1.45\pm0.22$  & 80\% \\
&P3 (n=13)& $\Theta^w = 80^{\circ} $  &  $1.40 \pm 0.14$ & 77\% \\
& Avg (n=33)&-& $\mathbf{1.62 \pm 0.41}$&$\mathbf{82\%}$\\
\bottomrule
\end{tabular}
\end{adjustbox}
\label{tbl:vs_table}
\end{table}
\subsection{Tissue Phantom Experiments } 
In this experiment, we first manually label a dissection goal on the image.
Our method will then select the optimal assistance position before controlling the forceps to satisfy the objectives. 
We repeated this process 33 times to assess the effectiveness and repeatability of the proposed method in enhancing visibility and creating tension for dissection tasks.


\begin{figure}[t]
    \centering
    \includegraphics[width=1\linewidth]{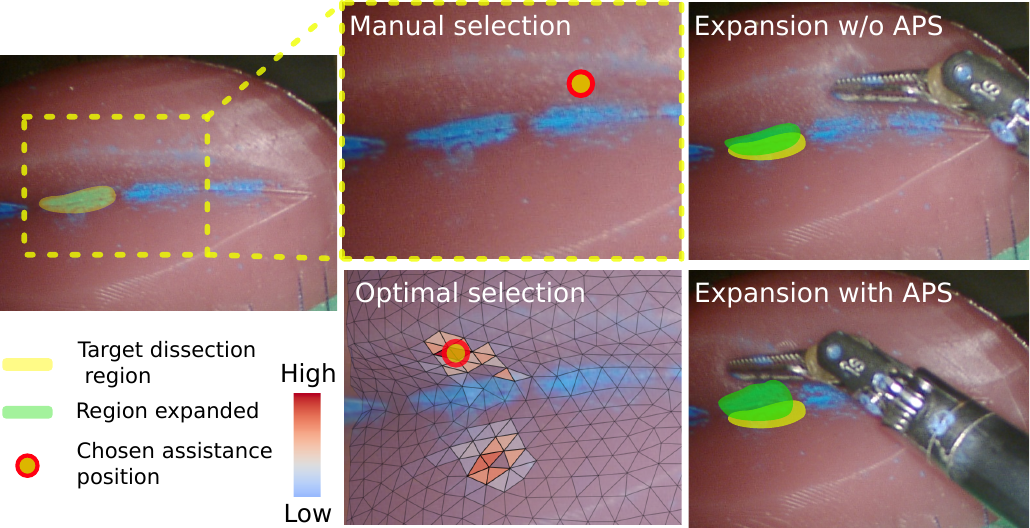}\vspace{-1.5em}
    \caption{A comparison between using our method with the proposed APS, and with a manually selected sub-optimal assistance position. The heatmap visualizes the computed $\mathcal{M}$ scores that APS uses to determine the optimal position. 
    Our method with APS achieves significantly more expansion of the dissection target than the suboptimal manual selection.}
    \label{fig:repositioning_analysis}\vspace{-1em}
    \vspace{-3mm}
\end{figure}

\textbf{Metric:} 
As visibility improvement are difficult to measure in real world scenarios, we propose a surrogate metric, $\rho$. 
We use a blue marker to paint dissection targets on the tissue phantom and measure the following ratio to quantify the effectiveness of the algorithm's control for dissection assistance: $\rho = \frac{\mathcal{A(\mathbf{x_{\textbf{T}}})}}{\mathcal{A(\mathbf{x_0})}}$, 
\begin{figure*}[t]
    \centering
    \includegraphics[width=1\linewidth]{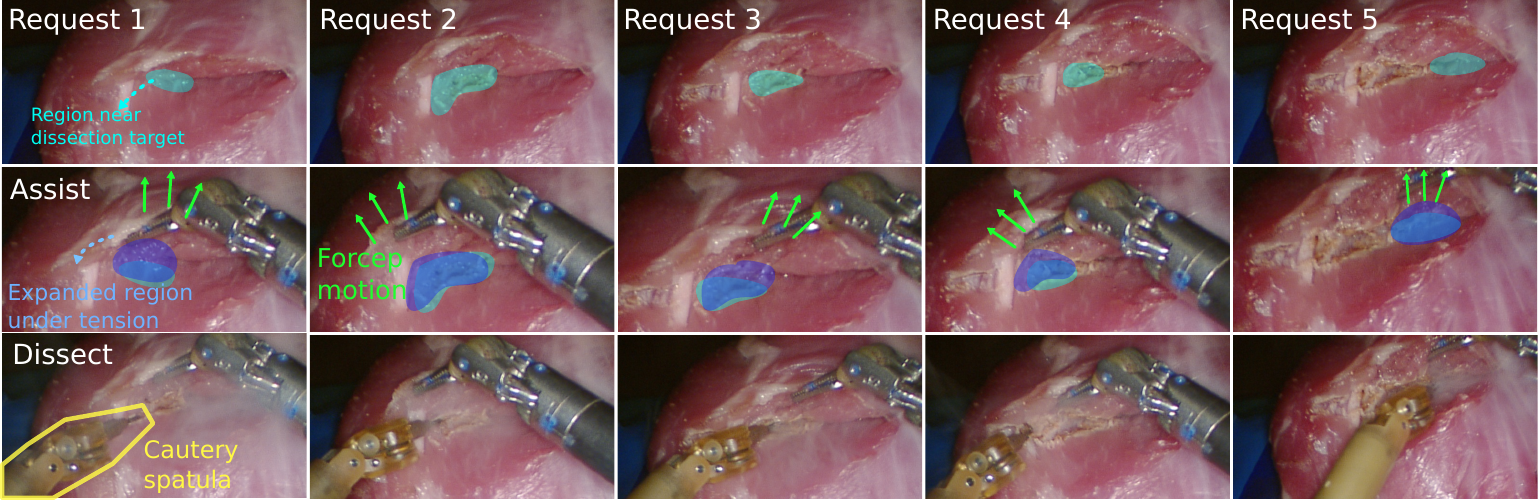}\vspace{-1.7em}
    \caption{This figure illustrates the execution of five consecutive  dissection assistance requests followed by manual dissection on real animal tissue.
    For each request, our method first locates the best assistance position and uses visual-servoing control to achieve the assistance objectives. 
    The human operator subsequently dissects the tissue with an electric cautery tool.
    The cyan regions near the dissection goals are expanded and tensioned to the larger blue region after application of our method.
    Between every request, we re-initalize the simulation to account for dissection effects.}
    \label{fig:real_experiments}
    \vspace{-1.5em}
\end{figure*}
which indicates the expansion ratio of dissection target in image after control at final time $t=\textbf{T}$. 
$\mathcal{A}(\textbf{x}_{t})$ is a measure of the blue pixels in the dissection target of the deformed tissue state, $\textbf{x}_{t}$, at time $t$. 
We measure this area by manually contouring out the blue dissection goal and computing its area. 
An example of the marked blue dissection target is shown in Fig.~\ref{fig:repositioning_analysis}. 
We consider the experiment a success if we see a visibility increase of $25\%$ or $\rho > 1.25$. 

\textbf{Effects of assistance position selection:} 
We analyze the effectiveness of proposed APS method.
We show a comparison of average final expansion ratio between our method with and without the proposed APS strategy in \autoref{fig:repositioning_analysis}.
When not using APS, we manually select a fixed assistance position that is around 1.5 cm distance away from the dissection target as shown in the figure.
The heat map shows that APS identifies regions that are close to and orthogonal to the dissection target which lead to a better control outcome.
We repeat this experiment 5 times using the same dissection target and the same fixed position. 
\autoref{tbl:aps_table} shows our method achieves a superior expansion ratio with APS, showcasing its positive impact on expanding the desired area. 

\textbf{Effects of visual servoing:}
To analyze the visual servoing controller, we compare our method to its variant (\textit{Ours-Hinge}) which treats the tissue surface as a hinge around the dissection line. 
This method plans an open loop trajectory to increase this hinge angle by forty-five degrees. 
We perform another 33 experiments on 3 phantoms with different initial opening angles and different dissection goals. 
Their results are summarized in Table.~\ref{tbl:vs_table}. With the  visual servoing controller, our method achieves an average expansion ratio of more than $1.6$ indicating that our method successfully increased exposure and improved visibility in the desired dissection region. In comparison, $\textit{ours-hinge}$ failed to do so due to naive assumption of the tissue dynamics, and lack of visual feedback. Our full method performs best when expanding smaller initial wedge angles, as they have the largest potential for increase in exposure. 
We achieve the highest success rate of $100\%$ on Phantom P1 with the smallest initial angle. 
Phantom P2 and P3 have larger wedge angles and, consequently, greater initial exposures, making it more difficult to achieve a substantial increase in exposure. 
In the failure cases on Phantom P2 and P3 our controller gets stuck in local minima, resulting in minimal movement. 
These failures can be attributed to inaccuracies in the jacobians due to modelling errors from the real to sim gap, as well as suboptimal parameter inputs such as poor selection of dissection targets and control objectives. 
\subsection{Real Tissue Experiments}
Finally, we deploy our method in a real tissue dissection scenario in which we apply our full method 5 times sequentially to assist dissecting the same tissue. 
Outcomes of this demo experiment are shown in Fig.\ref{fig:real_experiments}. For each dissection action, the human operator first labels an dissection target on the image (1st row). 
Next our algorithm computes the optimal assistance position from where it uses feedback control to deform the tissue using the forceps tool (2nd row). 
In the figure, the cyan regions near dissection targets are expanded to blue regions, improving visibility and applying tension to the dissection target. 
The human operator then performs dissection with the cautery spatula tool while the forceps tool stays at the final assistance position (3rd row). 
After every dissection action, we reconstruct mesh and restart simulation to proceed with the next request.

\subsection{Implementation Details}

We conducted experiments using $3$ ring radii $\mathbf{R} = \{r_1,r_2,r_3\}$ (specified in mm) and $5$ dissection segment points $\mathbf{L}=\{0, 0.25, 0.5, 0.75, 1\}$, yielding 15 feature pairs (30 points) in each experiment settings.  
We use ring radii $\textbf{R}_{phantoms} = \{3,6,9\}, \textbf{R}_{pork} = \{2,4,6\}$, stretch enforcement multipliers $\gamma^\textbf{v} = \gamma^\textbf{w} = 1.5$ and APS weighting parameter $\alpha_{phantom} = 0.1, \alpha_{pork} = 0.01$.
$l_{\text{min}}$ for APS is $0.005$.  
Proportional and derivative control gains are $K_p = 3 \times 10^{-4}$ and $K_d = 1 \times 10^{-5}$.
The maximum change in end-effector position per step was limited to 0.8 mm for phantom and 0.5 mm for pork tissue experiments to be realistic, with control step counts of 40 and 80, respectively.

\section{Discussion and Conclusion}


In this work, we propose MEDiC, a surgical robotic approach that assists a human operator to dissect tissue by manipulating the tissue to expose the dissection site.
We provided a complete framework for dissection assistance, from optimal control point selection to visual-servoing control. 
With our jacobian-informed approach, we are able to obtain an optimal control point from a complex tissue surface and manipulate the tissue to maximize the exposure of the dissection site and apply tension safely and efficiently.
By performing 38 trials on the phantoms and aiding in 5 consecutive dissections on real porcine tissue, we demonstrate the effectiveness of our method, and the promising future of shared autonomy in surgery.
Our method is even able to handle small variabilities in stiffness and irregular complex surface.
This is showcased in our experiments on real porcine tissue which shows natural variation in tissue attributes.   



In future works we seek to reduce the computational complexity required by our approach. 
We also seek to reduce our method's reliance on high fidelity models through more robust sensing techniques, to mitigate the issues caused by the real to sim gap.

\clearpage
\balance
\bibliographystyle{IEEEtran}
\bibliography{root}

\begin{thebibliography}{10}
\providecommand{\url}[1]{#1}
\csname url@samestyle\endcsname
\providecommand{\newblock}{\relax}
\providecommand{\bibinfo}[2]{#2}
\providecommand{\BIBentrySTDinterwordspacing}{\spaceskip=0pt\relax}
\providecommand{\BIBentryALTinterwordstretchfactor}{4}
\providecommand{\BIBentryALTinterwordspacing}{\spaceskip=\fontdimen2\font plus
\BIBentryALTinterwordstretchfactor\fontdimen3\font minus \fontdimen4\font\relax}
\providecommand{\BIBforeignlanguage}[2]{{%
\expandafter\ifx\csname l@#1\endcsname\relax
\typeout{** WARNING: IEEEtran.bst: No hyphenation pattern has been}%
\typeout{** loaded for the language `#1'. Using the pattern for}%
\typeout{** the default language instead.}%
\else
\language=\csname l@#1\endcsname
\fi
#2}}
\providecommand{\BIBdecl}{\relax}
\BIBdecl

\bibitem{dvrk}
P.~Kazanzides, Z.~Chen, A.~Deguet, G.~S. Fischer, R.~H. Taylor, and S.~P. DiMaio, ``An open-source research kit for the da vinci® surgical system,'' in \emph{2014 IEEE International Conference on Robotics and Automation (ICRA)}, 2014, pp. 6434--6439.

\bibitem{liang2023real}
X.~Liang, F.~Liu, Y.~Zhang, Y.~Li, S.~Lin, and M.~Yip, ``Real-to-sim deformable object manipulation: Optimizing physics models with residual mappings for robotic surgery,'' \emph{arXiv preprint arXiv:2309.11656}, 2023.

\bibitem{shinde2024jiggle}
N.~U. Shinde, X.~Liang, F.~Liu, Y.~Zhang, F.~Richter, S.~Herbert, and M.~C. Yip, ``Jiggle: An active sensing framework for boundary parameters estimation in deformable surgical environments,'' \emph{Conference on Robotics: Science and Systems (RSS)}, 2024.

\bibitem{Fei_2021_ICRA}
F.~Liu, Z.~Li, Y.~Han, J.~Lu, F.~Richter, and M.~C. Yip, ``Real-to-sim registration of deformable soft tissue with position-based dynamics for surgical robot autonomy,'' in \emph{2021 IEEE International Conference on Robotics and Automation (ICRA)}, 2021, pp. 12\,328--12\,334.

\bibitem{Jingbin_2021}
J.~Huang, F.~Liu, F.~Richter, and M.~C. Yip, ``Model-predictive control of blood suction for surgical hemostasis using differentiable fluid simulations,'' in \emph{2021 IEEE International Conference on Robotics and Automation (ICRA)}, 2021, pp. 12\,380--12\,386.

\bibitem{Florian_2021_RAL}
F.~Richter, S.~Shen, F.~Liu, J.~Huang, E.~K. Funk, R.~K. Orosco, and M.~C. Yip, ``Autonomous robotic suction to clear the surgical field for hemostasis using image-based blood flow detection,'' \emph{IEEE Robotics and Automation Letters}, vol.~6, no.~2, pp. 1383--1390, 2021.

\bibitem{dettorre2018automatedpickupsuturingneedles}
\BIBentryALTinterwordspacing
C.~D'Ettorre, G.~Dwyer, X.~Du, F.~Chadebecq, F.~Vasconcelos, E.~D. Momi, and D.~Stoyanov, ``Automated pick-up of suturing needles for robotic surgical assistance,'' 2018. [Online]. Available: \url{https://arxiv.org/abs/1804.03141}
\BIBentrySTDinterwordspacing

\bibitem{shinde2024surestep}
\BIBentryALTinterwordspacing
N.~U. Shinde, Z.-Y. Chiu, F.~Richter, J.~Lim, Y.~Zhi, S.~Herbert, and M.~C. Yip, ``Surestep: An uncertainty-aware trajectory optimization framework to enhance visual tool tracking for robust surgical automation,'' 2024. [Online]. Available: \url{https://arxiv.org/abs/2404.00123}
\BIBentrySTDinterwordspacing

\bibitem{chiu2023real}
Z.-Y. Chiu, F.~Richter, and M.~C. Yip, ``Real-time constrained 6d object-pose tracking of an in-hand suture needle for minimally invasive robotic surgery,'' in \emph{2023 IEEE International Conference on Robotics and Automation (ICRA)}.\hskip 1em plus 0.5em minus 0.4em\relax IEEE, 2023, pp. 4761--4767.

\bibitem{Neelay_2023_ICRA}
N.~Joglekar, F.~Liu, R.~Orosco, and M.~Yip, ``Suture thread spline reconstruction from endoscopic images for robotic surgery with reliability-driven keypoint detection,'' in \emph{2023 IEEE International Conference on Robotics and Automation (ICRA)}, 2023, pp. 4747--4753.

\bibitem{joglekar2024autonomousimagetograsproboticsuturing}
\BIBentryALTinterwordspacing
N.~Joglekar, F.~Liu, F.~Richter, and M.~C. Yip, ``Autonomous image-to-grasp robotic suturing using reliability-driven suture thread reconstruction,'' 2024. [Online]. Available: \url{https://arxiv.org/abs/2408.16938}
\BIBentrySTDinterwordspacing

\bibitem{oh2023framework}
K.-H. Oh, L.~Borgioli, M.~Zefran, L.~Chen, and P.~C. Giulianotti, ``A framework for automated dissection along tissue boundary,'' \emph{arXiv preprint arXiv:2310.09669}, 2023.

\bibitem{nguyen2019manipulating}
N.~D. Nguyen, T.~Nguyen, S.~Nahavandi, A.~Bhatti, and G.~Guest, ``Manipulating soft tissues by deep reinforcement learning for autonomous robotic surgery,'' in \emph{2019 IEEE International Systems Conference (SysCon)}.\hskip 1em plus 0.5em minus 0.4em\relax IEEE, 2019, pp. 1--7.

\bibitem{shahkoo2023autonomous}
A.~A. Shahkoo and A.~A. Abin, ``Autonomous tissue manipulation via surgical robot using deep reinforcement learning and evolutionary algorithm,'' \emph{IEEE Transactions on Medical Robotics and Bionics}, vol.~5, no.~1, pp. 30--41, 2023.

\bibitem{7989275}
B.~Thananjeyan, A.~Garg, S.~Krishnan, C.~Chen, L.~Miller, and K.~Goldberg, ``Multilateral surgical pattern cutting in 2d orthotropic gauze with deep reinforcement learning policies for tensioning,'' in \emph{2017 IEEE International Conference on Robotics and Automation (ICRA)}, 2017, pp. 2371--2378.

\bibitem{7139344}
A.~Murali, S.~Sen, B.~Kehoe, A.~Garg, S.~McFarland, S.~Patil, W.~D. Boyd, S.~Lim, P.~Abbeel, and K.~Goldberg, ``Learning by observation for surgical subtasks: Multilateral cutting of 3d viscoelastic and 2d orthotropic tissue phantoms,'' in \emph{2015 IEEE International Conference on Robotics and Automation (ICRA)}, 2015, pp. 1202--1209.

\bibitem{FundamentalsOfDissection}
\BIBentryALTinterwordspacing
N.~S. McCall and H.~Lavu, \emph{Fundamentals of Dissection}.\hskip 1em plus 0.5em minus 0.4em\relax Cham: Springer International Publishing, 2018, pp. 107--117. [Online]. Available: \url{https://doi.org/10.1007/978-3-319-75656-1_7}
\BIBentrySTDinterwordspacing

\bibitem{OperativeTechniquesCardiac}
D.~S. Hui, J.~M. Lizalek, V.~S. Chawa, and R.~Lee, ``Operative techniques for improving surgical exposure in basic cardiac surgery,'' \emph{J. Vis. Surg.}, vol.~4, pp. 80--80, Apr. 2018.

\bibitem{elek2017towards}
R.~Elek, T.~D. Nagy, D.~{\'A}. Nagy, T.~Garamv{\"o}lgyi, B.~Tak{\'a}cs, P.~Galambos, J.~K. Tar, I.~J. Rudas, and T.~Haidegger, ``Towards surgical subtask automation—blunt dissection,'' in \emph{2017 IEEE 21st International Conference on Intelligent Engineering Systems (INES)}.\hskip 1em plus 0.5em minus 0.4em\relax IEEE, 2017, pp. 000\,253--000\,258.

\bibitem{McKinley2015AutonomousMS}
\BIBentryALTinterwordspacing
S.~McKinley, A.~Garg, S.~Sen, D.~V. Gealy, J.~P. McKinley, Y.~Jen, and K.~Goldberg, ``Autonomous multilateral surgical tumor resection with interchangeable instrument mounts and fluid injection device,'' 2015. [Online]. Available: \url{https://api.semanticscholar.org/CorpusID:14696987}
\BIBentrySTDinterwordspacing

\bibitem{ge2021supervised}
J.~Ge, H.~Saeidi, M.~Kam, J.~Opfermann, and A.~Krieger, ``Supervised autonomous electrosurgery for soft tissue resection,'' in \emph{2021 IEEE 21st International Conference on Bioinformatics and Bioengineering (BIBE)}.\hskip 1em plus 0.5em minus 0.4em\relax IEEE, 2021, pp. 1--7.

\bibitem{10354422}
J.~Ge, M.~Kam, J.~D. Opfermann, H.~Saeidi, S.~Leonard, L.~J. Mady, M.~J. Schnermann, and A.~Krieger, ``Autonomous system for tumor resection (astr) - dual-arm robotic midline partial glossectomy,'' \emph{IEEE Robotics and Automation Letters}, vol.~9, no.~2, pp. 1166--1173, 2024.

\bibitem{8884158}
H.~Saeidi, J.~Ge, M.~Kam, J.~D. Opfermann, S.~Leonard, A.~S. Joshi, and A.~Krieger, ``Supervised autonomous electrosurgery via biocompatible near-infrared tissue tracking techniques,'' \emph{IEEE Transactions on Medical Robotics and Bionics}, vol.~1, no.~4, pp. 228--236, 2019.

\bibitem{10.1007/978-3-030-32239-7_36}
J.~Ge, H.~Saeidi, J.~D. Opfermann, A.~S. Joshi, and A.~Krieger, ``Landmark-guided deformable image registration for supervised autonomous robotic tumor resection,'' in \emph{Medical Image Computing and Computer Assisted Intervention -- MICCAI 2019}, D.~Shen, T.~Liu, T.~M. Peters, L.~H. Staib, C.~Essert, S.~Zhou, P.-T. Yap, and A.~Khan, Eds.\hskip 1em plus 0.5em minus 0.4em\relax Cham: Springer International Publishing, 2019, pp. 320--328.

\bibitem{karimi2024reward}
Z.~Karimi, S.-H. Ho, B.~Thach, A.~Kuntz, and D.~S. Brown, ``Reward learning from suboptimal demonstrations with applications in surgical electrocautery,'' \emph{arXiv preprint arXiv:2404.07185}, 2024.

\bibitem{marahrens2024ultrasound}
N.~Marahrens, D.~Jones, N.~Murasovs, C.~Biyani, and P.~Valdastri, ``An ultrasound-guided system for autonomous marking of tumor boundaries during robot-assisted surgery,'' \emph{IEEE Transactions on Medical Robotics and Bionics}, 2024.

\bibitem{franccois2021image}
T.~Fran{\c{c}}ois, L.~Calvet, C.~S{\`e}ve-d’Erceville, N.~Bourdel, and A.~Bartoli, ``Image-based incision detection for topological intraoperative 3d model update in augmented reality assisted laparoscopic surgery,'' in \emph{Medical Image Computing and Computer Assisted Intervention--MICCAI 2021: 24th International Conference, Strasbourg, France, September 27--October 1, 2021, Proceedings, Part IV 24}.\hskip 1em plus 0.5em minus 0.4em\relax Springer, 2021, pp. 647--656.

\bibitem{DBLP:journals/corr/abs-1901-03327}
\BIBentryALTinterwordspacing
T.~T. Nguyen, N.~D. Nguyen, F.~Bello, and S.~Nahavandi, ``A new tensioning method using deep reinforcement learning for surgical pattern cutting,'' \emph{CoRR}, vol. abs/1901.03327, 2019. [Online]. Available: \url{http://arxiv.org/abs/1901.03327}
\BIBentrySTDinterwordspacing

\bibitem{5354075}
R.~Jansen, K.~Hauser, N.~Chentanez, F.~van~der Stappen, and K.~Goldberg, ``Surgical retraction of non-uniform deformable layers of tissue: 2d robot grasping and path planning,'' in \emph{2009 IEEE/RSJ International Conference on Intelligent Robots and Systems}, 2009, pp. 4092--4097.

\bibitem{patil2010toward}
S.~Patil and R.~Alterovitz, ``Toward automated tissue retraction in robot-assisted surgery,'' in \emph{2010 IEEE International Conference on Robotics and Automation}.\hskip 1em plus 0.5em minus 0.4em\relax IEEE, 2010, pp. 2088--2094.

\bibitem{meli2021autonomous}
D.~Meli, E.~Tagliabue, D.~Dall’Alba, and P.~Fiorini, ``Autonomous tissue retraction with a biomechanically informed logic based framework,'' in \emph{2021 International Symposium on Medical Robotics (ISMR)}.\hskip 1em plus 0.5em minus 0.4em\relax IEEE, 2021, pp. 1--7.

\bibitem{li2022practical}
J.~Li, P.~Wang, P.~Xiong, T.~Cai, Z.~Yan, L.~Yang, J.~Liu, H.~Fan, and S.~Liu, ``Practical stereo matching via cascaded recurrent network with adaptive correlation,'' in \emph{Proceedings of the IEEE/CVF conference on computer vision and pattern recognition}, 2022, pp. 16\,263--16\,272.

\bibitem{SAM}
A.~Kirillov, E.~Mintun, N.~Ravi, H.~Mao, C.~Rolland, L.~Gustafson, T.~Xiao, S.~Whitehead, A.~C. Berg, W.-Y. Lo, P.~Doll{\'a}r, and R.~Girshick, ``Segment anything,'' \emph{arXiv:2304.02643}, 2023.

\bibitem{cheng2024putting}
H.~K. Cheng, S.~W. Oh, B.~Price, J.-Y. Lee, and A.~Schwing, ``Putting the object back into video object segmentation,'' in \emph{Proceedings of the IEEE/CVF Conference on Computer Vision and Pattern Recognition}, 2024, pp. 3151--3161.

\end{thebibliography}

\end{document}